\documentclass[]{template/mac_automl_xmu_blue}

\graphicspath{{content/}}

\usepackage[toc,page,header]{appendix}

\usepackage{amsthm}
\usepackage{graphicx}
\usepackage{caption}
\usepackage{placeins}

\newcommand{\equalcontribmark}{\textsuperscript{*}}

\usepackage{xargs}
\usepackage{todonotes}
\usepackage{amsmath,amsfonts,amssymb,bm}
\usepackage{graphicx}
\usepackage{hyperref}
\usepackage{url}
\usepackage{booktabs}
\usepackage{tabularx}
\usepackage{multirow}
\usepackage{colortbl}
\definecolor{lightbluebar}{RGB}{232,244,255}

\def\1{\bm{1}}

\setleftheadercontent{%
  \headerlogospace{1.6mm}%
  \headerlogo[-0.08\height]{15.2mm}{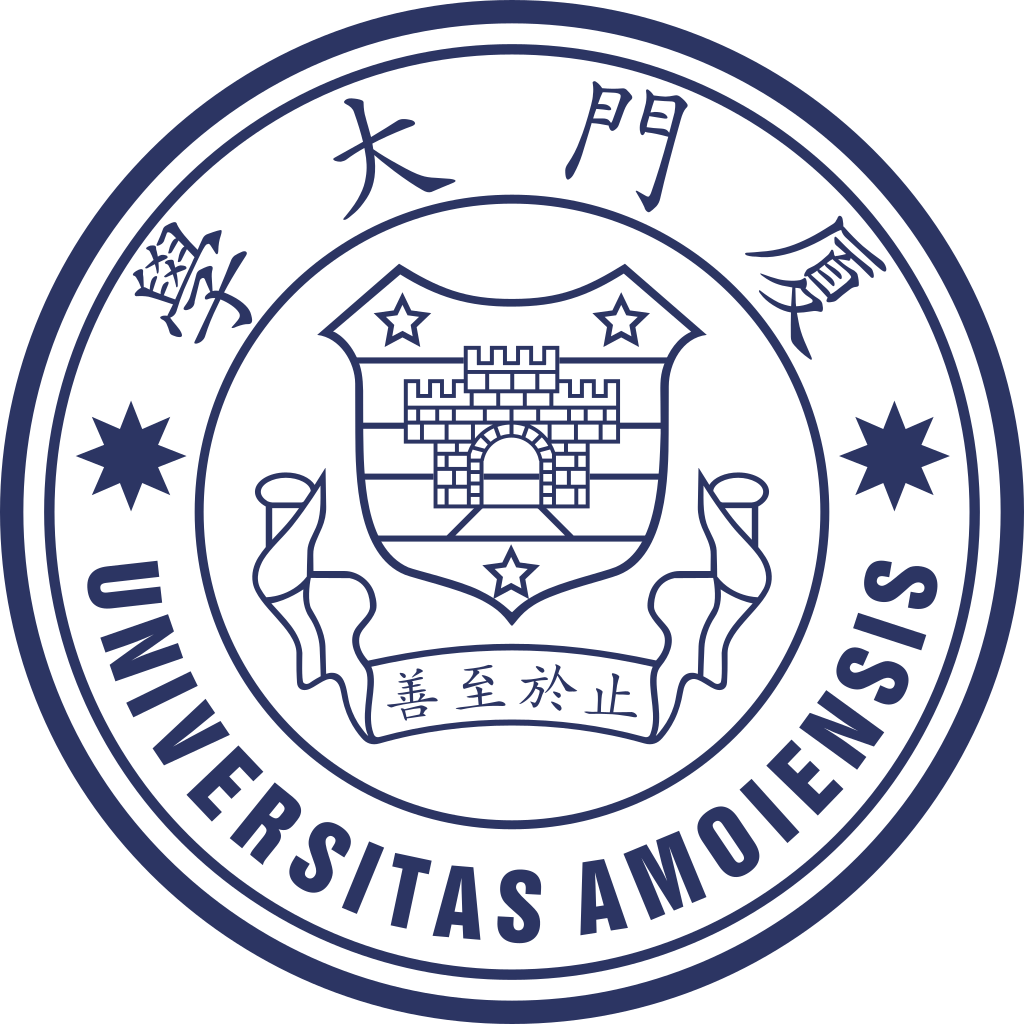}%
}
\setrightheadericon{
  \headerlogospace{1.6mm}%
  \headerlogo[-0.08\height]{15.2mm}{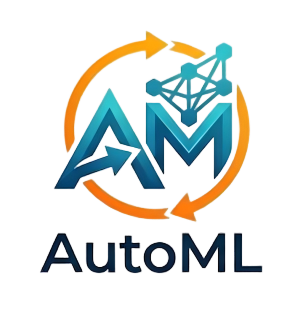}%
}
\setheadergroupname{MAC-AutoML}

\title{HEART: Exploiting Head Heterogeneity in Sparse Attention for Video Diffusion}

\setfrontauthors{%
  \authorrow{%
    \authorentry{Xuzhe Zheng}{1}\equalcontribmark,\authorsep
    \authorentry{Yuexiao Ma}{1}\equalcontribmark,\authorsep
    \authorentry{Jing Xu}{1},\authorsep
    \authorentry{Xiawu Zheng}{1},\authorsep
    \authorentry{Rongrong Ji}{1},\authorsep
    \authorentry{Fei Chao}{1,\corremailmark}}%
}
\setfrontaffiliations{%
  \affiliationline{\authormark{1}Key Laboratory of Multimedia Trusted Perception and Efficient Computing, Ministry of Education of China, Xiamen University, 361005, P.R. China}
}
\setfrontcontact{%
\contactline{\equalcontribmark\ Equal contribution}
\contactline{\textbf{\corremailmark\ Corresponding Author}}
}
\usecustomauthorlayout

\abstract{Sparse attention accelerates video diffusion by allowing each attention head to focus on only a small subset of interactions. Existing methods already construct head-specific sparse patterns conditioned on the input. However, we find that these heads also differ in two less obvious but practically important ways. First, for some heads, sparse attention masks remain stable across many denoising steps, whereas others change rapidly. Second, heads differ substantially in their sensitivity to sparsification: applying the same threshold can induce markedly different errors in the final denoising velocity. Ignoring these differences leads to redundant mask prediction and suboptimal threshold calibration across heads under a global sparsity budget. We present HEART, short for Heterogeneity-Exploiting Adaptive Refresh and Thresholding, a training-free framework that exploits both forms of head heterogeneity. First, Temporal Mask Reuse (TMR) uses a lightweight per-head query-key drift signal to determine whether a cached sparse mask remains reliable across denoising steps, refreshing it only when the drift exceeds a prescribed threshold. Second, Error-guided Budgeted Calibration (EBC) evaluates candidate thresholds offline using a frequency-weighted denoising-velocity error, and assigns each head an appropriate threshold under a global sparsity budget. HEART requires no retraining, weight modification, or sparse-kernel changes, and can be integrated directly into existing sparse-attention pipelines. Across Wan2.1-1.3B, Wan2.1-14B, and HunyuanVideo-13B, HEART consistently pushes the quality--efficiency frontier of advanced sparse attention methods such as XAttention and SVG2 outward.
}

\begin{document}

\maketitle

\section{Introduction}

Video diffusion transformers \citep{peebles2023scalable,ma2024latte,ma2025step,yang2025cogvideox,kong2024hunyuanvideo,wan2025wan} have become a foundation of modern video generation, driving major advances in visual fidelity, temporal coherence, and long-horizon synthesis. As these models move toward higher resolutions and longer videos, improving their inference efficiency becomes increasingly important for practical deployment. Video diffusion transformers process videos as long spatiotemporal token sequences, with both spatial resolution and temporal length contributing to the total token count \citep{team2025longcat,zheng2025open,wu2025hunyuanvideo}. As the sequence grows, the quadratic cost of full attention makes it a major runtime bottleneck in DiTs \citep{wen2025analysis}. Training-free sparse attention directly addresses this issue by retaining only a subset of attention interactions at inference time, thereby reducing the attention cost of each denoising step without retraining or modifying pretrained weights \citep{chen2026sparse,li2026radial,xia2025training}.

\begin{figure*}[!t]
\centering
\includegraphics[width=\textwidth]{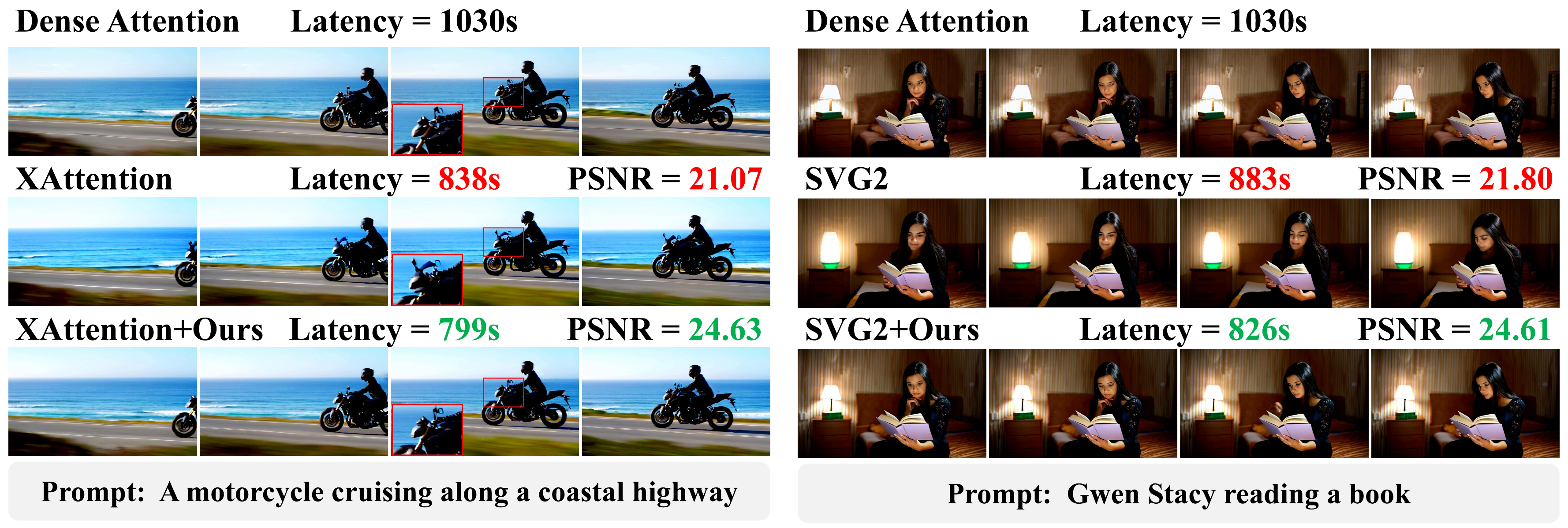}
\caption{
Qualitative comparison on Wan2.1-14B text-to-video generation at 480P. We compare dense attention with two representative sparse-attention baselines, XAttention \citep{xu2025xattention} and SVG2 \citep{yang2026sparse}, and their variants enhanced by our method.
These results show that our method better preserves the video quality while further improving inference efficiency.
}
\label{fig:qualitative_comparison_14b_480p}
\end{figure*}

Recent training-free sparse attention methods for video generation increasingly adopt online top-$p$ sparsification \citep{gu2025blade,yang2026sparse,wu2025vmoba}, which retains attention blocks until a cumulative importance mass is reached rather than keeping a fixed number of blocks. Compared with fixed sparse patterns \citep{li2026radial,chen2026sparse,zhao2026paroattention}, online methods are more flexible, but they still suffer from two practical inefficiencies. The first is mask-prediction overhead: the sparse mask of each head is typically re-predicted at every denoising step of the diffusion process, leading to non-negligible computational overhead. The second is shared-threshold suboptimality: existing methods usually apply one shared threshold across all heads, despite the fact that different heads can differ substantially in how strongly sparsification errors propagate to the final denoising velocity. As a result, current methods can be suboptimal both in when to refresh sparse masks and in how to allocate head-specific thresholds under a global sparsity budget.

We address these two issues with a head-wise adaptive sparse-attention framework. The first component, \textbf{Temporal Mask Reuse (TMR)}, reduces online mask-prediction overhead by deciding, for each head and denoising step, whether the cached sparse mask can still be reused or needs to be refreshed. The decision is based on a lightweight query-key drift statistic, so the reuse frequency adapts to the actual evolution of each head. The second component, \textbf{Error-guided Budgeted Calibration (EBC)}, assigns head-specific thresholds under a global sparsity budget by minimizing a frequency-aware velocity error, so different heads can be calibrated according to their heterogeneous impact on the final denoising velocity. TMR and EBC are complementary: the former reduces online overhead, while the latter improves threshold allocation across heterogeneous heads under a shared sparsity budget.

Both components are plug-in modifications to existing sparse-attention pipelines and require neither retraining nor changes to the underlying sparse kernels. Across Wan2.1-1.3B, Wan2.1-14B, and HunyuanVideo-13B \citep{wan2025wan,kong2024hunyuanvideo}, our method consistently improves the quality--efficiency frontier of advanced sparse attention methods such as XAttention \citep{xu2025xattention} and SVG2 \citep{yang2026sparse}, showing that the benefit is not tied to a specific sparse-attention design. For example, on Wan2.1-1.3B at 720P, our method improves XAttention from $1.67\times$ to $1.85\times$ end-to-end speedup and from $2.23\times$ to $2.68\times$ attention-only speedup, while also improving both video quality and similarity metrics. We observe the same outward shift of the quality--efficiency frontier on larger and different backbones, including Wan2.1-14B and HunyuanVideo-13B, confirming that head-wise adaptive reuse and error-aware calibration generalize across model scales, resolutions, and sparse-attention backbones.

Our main contributions are summarized as follows:

\begin{enumerate}
\item We identify two under-explored forms of head-wise heterogeneity: temporal mask-evolution heterogeneity across denoising steps and heterogeneity in how sparsification errors from different heads affect the final denoising velocity. These observations reveal why fixed mask-refresh schedules and shared thresholds are suboptimal for practical speed--quality trade-offs.

\item Motivated by these observations, we propose a head-wise adaptive sparse-attention framework with two complementary components: Temporal Mask Reuse (TMR), which reduces online mask prediction overhead through adaptive mask reuse across denoising steps, and Error-guided Budgeted Calibration (EBC), which selects head-specific thresholds according to a frequency-aware velocity error rather than relying on a shared threshold. The framework is training-free and can be integrated into existing sparse-attention pipelines without modifying pretrained weights or sparse kernels.

\item We instantiate the proposed framework on representative sparse-attention baselines, including XAttention \citep{xu2025xattention} and SVG2 \citep{yang2026sparse}, and conduct extensive experiments on Wan2.1-1.3B, Wan2.1-14B, and HunyuanVideo-13B \citep{wan2025wan,kong2024hunyuanvideo}. Results show consistent improvements in the speed--quality trade-off across model scales, resolutions, and sparse-attention backbones.

\end{enumerate}

\section{Related Work}

SVG identifies spatial and temporal heads through online profiling and exploits their distinct sparse patterns with hardware-aware layouts and kernels \citep{xi2025sparse}. SVG2 uses semantic-aware token permutation to improve critical-token identification and pack retained tokens into a more GPU-efficient layout \citep{yang2026sparse}. XAttention uses antidiagonal scoring as a low-cost proxy for block importance \citep{xu2025xattention}. SpargeAttn applies a two-stage online filtering scheme to skip redundant attention computation \citep{zhang2025spargeattention}.

More closely related to our work are methods that study \emph{when} to refresh sparse masks and \emph{how} to allocate head-specific thresholds under sparsity constraints. On the mask-prediction side, LiteAttention, AdaSpa, and PAROAttention exploit temporal coherence by reusing previously constructed masks across denoising steps, thereby avoiding mask reconstruction at every step \citep{shmilovich2025liteattention,xia2025training,zhao2026paroattention}. On the threshold-allocation side, recent methods profile heterogeneity over layers, heads, or timesteps and distribute computation more selectively \citep{zhang2026ride,zhan2025bidirectional,wu2025vmoba}. HEART combines these two directions: it makes mask-refresh decisions head-wise and adaptive at runtime, while calibrating head-specific thresholds under a global sparsity budget using velocity-aware offline measurements.

\section{Key Observations}

\subsection{Observation I: Head-Wise Heterogeneity in Mask Evolution}
\label{sec:obs1}

\begin{figure}[t]
\centering
\includegraphics[width=\columnwidth]{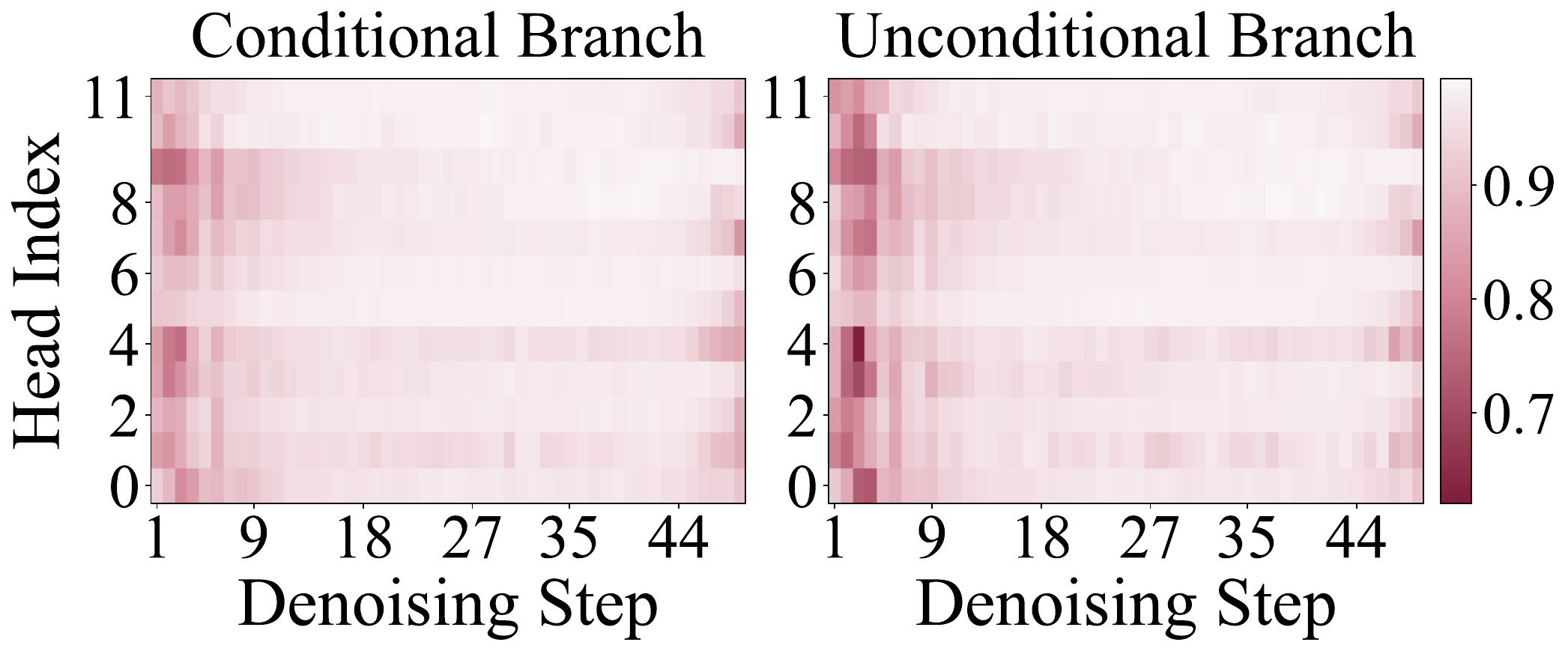}
\caption{Temporal mask similarity for heads from a randomly selected layer, measured by adjacent-step mask IoU. Different heads follow noticeably different temporal trajectories, indicating that cross-step mask persistence varies substantially across heads.}
\label{fig:temporal-mask-similarity}
\end{figure}

In online sparse attention, sparse patterns are not predetermined but dynamically predicted during inference, introducing additional overhead \citep{xu2025xattention,yang2026sparse,xia2025training,chen2025rainfusion2}. This raises a natural question: must the sparse mask be recomputed at every denoising step, or can a recently predicted mask be reused safely?

To study this, we measure how sparse masks evolve across adjacent denoising steps. For head \(h\), let \(A_t^{(h)}\) denote the post-softmax attention matrix at denoising step \(t\). We define the corresponding binary attention mask as
\begin{equation}
M_t^{(h)}(i,j) = \mathbf{1}\!\left[j \in \mathcal{T}_{0.95}\!\bigl(A_t^{(h)}(i,:)\bigr)\right]
\label{eq:binary_mask}
\end{equation}
where $\mathcal{T}_{0.95}(\cdot)$ returns the minimal set of key indices whose cumulative probability mass reaches $0.95$. Adjacent-step mask similarity is then measured by
\begin{equation}
\operatorname{IoU}_t^{(h)}
=
\frac{|M_t^{(h)} \cap M_{t+1}^{(h)}|}{|M_t^{(h)} \cup M_{t+1}^{(h)}|}.
\end{equation}

Figure~\ref{fig:temporal-mask-similarity} visualizes this quantity for heads from a randomly selected layer. The trajectories vary substantially across heads: some remain relatively stable over multiple denoising steps, while others change much more rapidly.

This observation suggests that sparse masks need not be re-predicted at every denoising step, because for some heads at some steps the sparse mask remains relatively stable. However, predefined reuse schedules, such as fixed-interval refresh or switching to reuse after a preset denoising step, are still too rigid to capture this head-wise and step-wise variation. Mask reuse should instead be controlled at head granularity according to the actual evolution of each head.

\subsection{Observation II: Head-Wise Heterogeneity in Top-$p$ Threshold Responses}
\label{sec:obs2}

In top-$p$ sparse attention, each head is controlled by a cumulative-mass threshold \(\tau\): candidate blocks are ranked by importance and retained until their accumulated importance reaches \(\tau\) \citep{yang2026sparse,zhang2025spargeattention}. Unlike top-$k$, where the number of retained blocks is fixed directly \citep{hu2026dfsattn,li2026pisa,tan2026adacluster}, top-$p$ does not prescribe a fixed sparsity level. The realized sparsity instead depends on the score distribution of the head itself: flatter distributions require more blocks to reach the same cumulative mass, whereas more peaked distributions require fewer. As a result, applying the same threshold to different heads can produce substantially different realized sparsity levels.

Beyond realized sparsity, the error introduced by sparse attention can also differ across heads, and it can be measured at two levels: the local attention output and the denoising velocity used for sampling. These two are not equivalent. A perturbation introduced at one attention head is further transformed by downstream layers before it affects the denoising velocity, so a threshold that appears acceptable under attention-output error may still produce a disproportionate error in the denoising velocity. The quantity that ultimately matters is therefore the error on denoising velocity rather than a purely local attention-output error.

\begin{figure*}[!t]
\centering
\includegraphics[width=\textwidth]{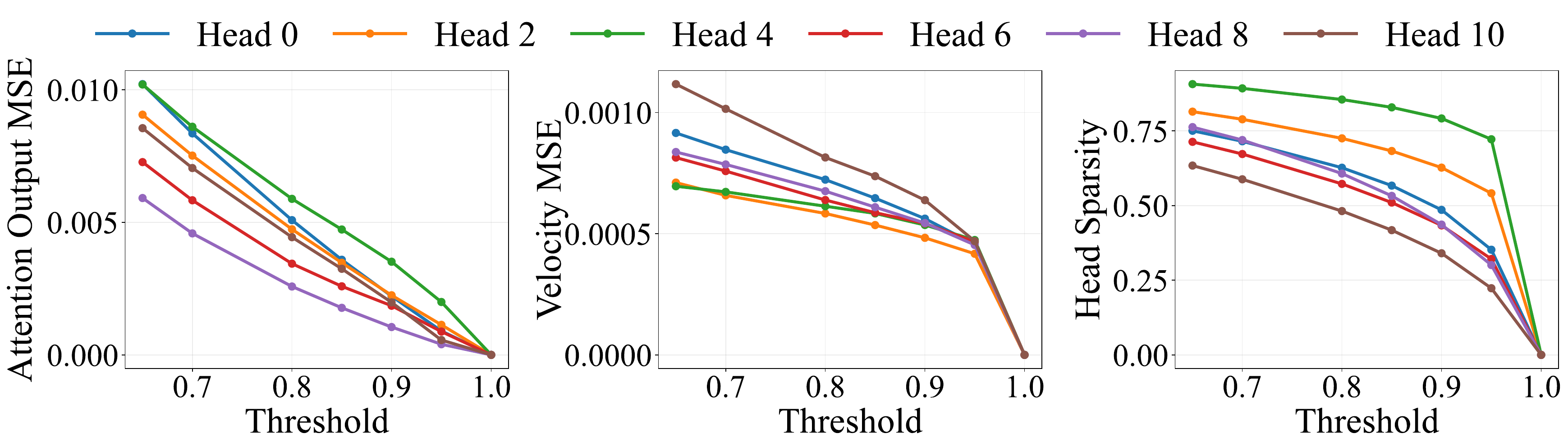}
\caption{Head-wise effects of top-$p$ thresholds under XAttention on Wan2.1-1.3B at 480P. Each curve represents one head: left, sparse-attention-induced attention-output error; middle, induced denoising-velocity error; right, realized sparsity. The diverse head-wise responses motivate head-specific threshold calibration.}
\label{fig:head-tradeoff-layer15}
\end{figure*}

Figure~\ref{fig:head-tradeoff-layer15} shows that different heads respond very differently to the same threshold. Under an identical threshold, some heads can already achieve relatively high sparsity while keeping the sparse-attention-induced error small, whereas others exhibit much larger error or much lower sparsity. Therefore, a shared threshold does not place different heads at comparable sparsity--error regimes.

If the goal is to keep the effect of sparsification on the denoising velocity similarly small across heads, then different heads should generally use different thresholds. This directly motivates our error-guided budgeted calibration method.

\section{Method}

\subsection{Overview}

\begin{figure*}[t]
\centering
\includegraphics[width=\linewidth]{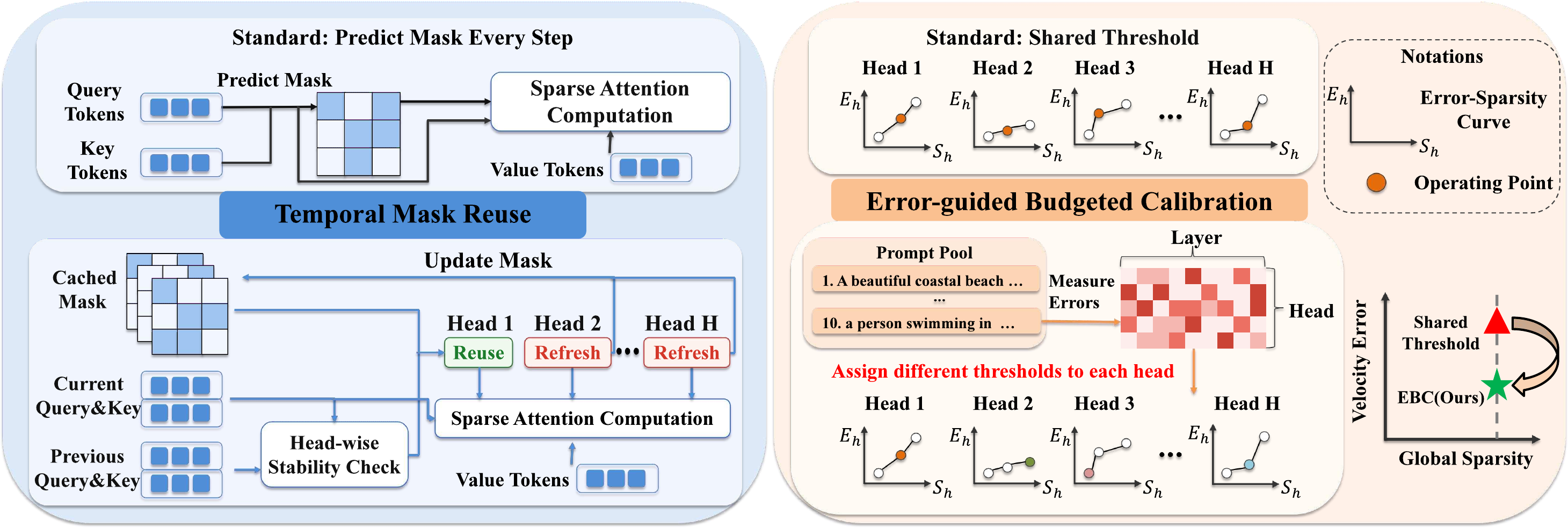}
\caption{Overview of HEART, which enhances existing sparse attention with two complementary head-wise mechanisms. Left: Temporal Mask Reuse (TMR) operates online during denoising by comparing current and previous query-key features for each head. Stable heads reuse their cached sparse masks, while unstable heads refresh their masks before sparse attention computation. Right: Error-guided Budgeted Calibration (EBC) operates offline. Each candidate operating point pairs a threshold with its measured sparsity and velocity error. EBC replaces the shared threshold used by conventional sparse attention with head-specific thresholds calibrated from these measurements, better preserving denoising velocity at the same target sparsity.}
\label{fig:method-overview}
\end{figure*}

Figure~\ref{fig:method-overview} gives an overview of the proposed framework.  The framework consists of two complementary modules designed to augment existing sparse-attention pipelines.

\textbf{Temporal Mask Reuse (TMR)} decides, for each head, whether to reuse the mask from the most recent refresh step or to predict a new one at the current step. It operates online through a lightweight query-key temporal stability statistic and is applied during the mask computation stage of an existing pipeline.

\textbf{Error-guided Budgeted Calibration (EBC)} determines the top-$p$ threshold of each attention head. It operates offline by measuring candidate operating points, where each operating point consists of a threshold together with its resulting sparsity and velocity error, and then selecting one calibrated threshold per head under a global sparsity budget, while leaving the underlying pipeline unchanged.

\subsection{Temporal Mask Reuse}

Observation~\ref{sec:obs1} shows that mask stability varies substantially across heads and denoising steps, making a shared reuse schedule too coarse. TMR therefore makes the refresh decision at head granularity. For each head \(h\), it maintains an anchor step \(t_a\), i.e., the most recent denoising step at which the sparse mask was refreshed. At a later denoising step \(t_b>t_a\), TMR decides whether the cached mask from \(t_a\) can still be reused or whether a new mask should be predicted.

The main challenge is that mask stability cannot be checked by explicitly reconstructing the current sparse mask, because doing so would defeat the purpose of reuse. We therefore use query-key drift as a proxy for cross-step mask change:
\begin{equation}
d_{t_a\rightarrow t_b}^{(h)}
:=
\frac{1}{N}\sum_{i=1}^N
\left(
\left\|q_{t_a,i}^{(h)}-q_{t_b,i}^{(h)}\right\|_1
+
\left\|k_{t_a,i}^{(h)}-k_{t_b,i}^{(h)}\right\|_1
\right) .
\label{eq:full_token_drift}
\end{equation}
This proxy is motivated by how practical sparse attention is executed. The sparse pattern is determined by block-level importance ranking \citep{xu2025xattention,zhang2025spargeattention}. When the query and key features of a head change only mildly across denoising steps, the induced block-importance ranking is also less likely to change substantially, so the cached sparse mask is more likely to remain valid.

Using Eq.~\eqref{eq:full_token_drift} directly is still too expensive because it requires token-level caching. We therefore use the following mean-pooled approximation online:
\begin{equation}
\tilde d_{t_a\rightarrow t_b}^{(h)}
=
\left\|\bar Q_{t_a}^{(h)}-\bar Q_{t_b}^{(h)}\right\|_1
+
\left\|\bar K_{t_a}^{(h)}-\bar K_{t_b}^{(h)}\right\|_1,
\label{eq:mean_pool_drift}
\end{equation}
where \(\bar Q_t^{(h)}\) and \(\bar K_t^{(h)}\) are token-averaged query and key features. This reduces cache from token-level features to one pooled query and key vector per head. In Wan2.1-1.3B 480P generation with classifier-free guidance, storing full-token query/key features in FP16 requires about \(11.2\) GB under \(L{=}30\), \(H{=}12\), \(D{=}128\), and \(N{=}32{,}760\), whereas the mean-pooled cache requires only about \(0.35\) MB, making online control practical.

This approximation remains useful empirically. The full-token drift and the mean-pooled proxy have Spearman correlations of \(-0.7914\) and \(-0.7568\) with adjacent-step mask IoU, respectively, indicating that the cheaper token-mean proxy remains effective for reuse decisions while avoiding the cost of token-level caching.

The final online rule is simple. If \(\tilde d_{t_a\rightarrow t_b}^{(h)} \le \delta\), head \(h\) reuses its cached mask from the anchor step; otherwise the mask is refreshed and the anchor is updated to \(t_b\). TMR therefore adds only lightweight per-head pooling and \(L_1\) comparisons, while allowing refresh decisions to vary across heads and denoising steps.

Although TMR decides mask refresh at head granularity, the actual cost of sparse-mask construction is not proportional to the number of refreshed heads. In practice, sparse-mask construction within a layer incurs non-negligible fixed runtime overhead, such as kernel launches, even when only a small subset of heads is refreshed. As a result, refreshing only a few heads may save less time than the refreshed-head count alone would suggest, while refreshing most heads separately can also be inefficient. We therefore add a lightweight layer-level gating heuristic based on the fraction of heads marked for refresh in each layer. If this fraction is below a lower threshold, the entire layer reuses cached masks; if it exceeds an upper threshold, the entire layer is refreshed; otherwise, the original head-wise TMR decisions are applied.

\begin{table}[b]
\centering
{\small
\begin{tabular}{@{}lccc@{}}
\toprule
Region & PSNR$\uparrow$ & SSIM$\uparrow$ & LPIPS$\downarrow$ \\
\midrule
LL & 28.19 & 0.8452 & 0.1024 \\
LH & 29.58 & 0.8485 & 0.0970 \\
HL & 29.85 & 0.8529 & 0.0818 \\
HH & 30.35 & 0.8634 & 0.0791 \\
\bottomrule
\end{tabular}
}
\caption{Sensitivity of generated-video quality to equal-magnitude perturbations in different 3D-FFT regions of the dense denoising velocity. Lower-frequency perturbations are more harmful, motivating frequency-aware calibration.}
\label{tab:frequency-sensitivity}
\end{table}

\begin{table*}[t]
\centering
{\small
\setlength{\tabcolsep}{1mm}
\begin{tabular}{@{}llccccccccc@{}}
\toprule
\multirow{2}{*}{Model} & \multirow{2}{*}{Method} & \multicolumn{2}{c}{Quality} & \multicolumn{3}{c}{Similarity} & \multicolumn{4}{c}{Efficiency} \\
\cmidrule(lr){3-4}\cmidrule(lr){5-7}\cmidrule(lr){8-11}
& & VBench$\uparrow$ & VReward$\uparrow$ & PSNR$\uparrow$ & SSIM$\uparrow$ & LPIPS$\downarrow$ & Attn Lat.$\downarrow$ & Attn Spd.$\uparrow$ & E2E Lat.$\downarrow$ & E2E Spd.$\uparrow$ \\
\midrule
\multirow{5}{*}{Wan2.1-1.3B} & Dense & 77.15\% & 0.0862 & -- & -- & -- & 107s & 1.00$\times$ & 195s & 1.00$\times$ \\
& XAttention & 76.99\% & 0.0535 & 21.04 & 0.6481 & 0.2527 & 69s & 1.55$\times$ & 154s & 1.27$\times$ \\
& \cellcolor{lightbluebar}XAttn+Ours & \cellcolor{lightbluebar}\textbf{77.04\%} & \cellcolor{lightbluebar}\textbf{0.0585} & \cellcolor{lightbluebar}\textbf{21.61} & \cellcolor{lightbluebar}\textbf{0.6833} & \cellcolor{lightbluebar}\textbf{0.2337} & \cellcolor{lightbluebar}\textbf{48s} & \cellcolor{lightbluebar}\textbf{2.23$\times$} & \cellcolor{lightbluebar}\textbf{135s} & \cellcolor{lightbluebar}\textbf{1.44$\times$} \\
& SVG2 & 76.85\% & 0.0678 & 23.84 & 0.7569 & 0.1678 & 88s & 1.22$\times$ & 156s & 1.25$\times$ \\
& \cellcolor{lightbluebar}SVG2+Ours & \cellcolor{lightbluebar}\textbf{76.89\%} & \cellcolor{lightbluebar}\textbf{0.0698} & \cellcolor{lightbluebar}\textbf{24.78} & \cellcolor{lightbluebar}\textbf{0.7775} & \cellcolor{lightbluebar}\textbf{0.1460} & \cellcolor{lightbluebar}\textbf{72}s & \cellcolor{lightbluebar}\textbf{1.49}$\times$ & \cellcolor{lightbluebar}\textbf{140s} & \cellcolor{lightbluebar}\textbf{1.39$\times$} \\
\midrule
\multirow{5}{*}{\shortstack{Wan2.1-1.3B\\(720P)}} & Dense & {75.52}\% & {0.0821} & -- & -- & -- & {542}s & {1.00}$\times$ & {740}s & 1.00$\times$ \\
& XAttention & {75.08}\% & {0.0622} & 22.24 & 0.7076 & 0.2876 & {243}s & {2.23}$\times$ & {442}s & 1.67$\times$ \\
& \cellcolor{lightbluebar}XAttn+Ours & \cellcolor{lightbluebar}\textbf{75.19}\% & \cellcolor{lightbluebar}\textbf{0.0673} & \cellcolor{lightbluebar}\textbf{22.69} & \cellcolor{lightbluebar}\textbf{0.7320} & \cellcolor{lightbluebar}\textbf{0.2547} & \cellcolor{lightbluebar}\textbf{202}s & \cellcolor{lightbluebar}\textbf{2.68}$\times$ & \cellcolor{lightbluebar}\textbf{401}s & \cellcolor{lightbluebar}\textbf{1.85$\times$} \\
& SVG2 & {74.93}\% & {0.0710} & 26.00 & 0.8414 & 0.1184 & {272}s & {1.99}$\times$ & {442}s & 1.67$\times$ \\
& \cellcolor{lightbluebar}SVG2+Ours & \cellcolor{lightbluebar}\textbf{74.87}\% & \cellcolor{lightbluebar}\textbf{0.0729} & \cellcolor{lightbluebar}\textbf{26.07} & \cellcolor{lightbluebar}\textbf{0.8442} & \cellcolor{lightbluebar}\textbf{0.1134} & \cellcolor{lightbluebar}\textbf{240}s & \cellcolor{lightbluebar}\textbf{2.26}$\times$ & \cellcolor{lightbluebar}\textbf{406}s & \cellcolor{lightbluebar}\textbf{1.82$\times$} \\
\midrule
\multirow{5}{*}{Wan2.1-14B} & Dense & 79.31\% & 0.0927 & -- & -- & -- & {461}s & {1.00}$\times$ & 1030s & 1.00$\times$ \\
& XAttention & 78.74\% & 0.0857 & 21.07 & 0.6959 & 0.2236 & {302}s & {1.53}$\times$ & 838s & 1.23$\times$ \\
& \cellcolor{lightbluebar}XAttn+Ours & \cellcolor{lightbluebar}\textbf{78.88\%} & \cellcolor{lightbluebar}\textbf{0.0863} & \cellcolor{lightbluebar}\textbf{24.63} & \cellcolor{lightbluebar}\textbf{0.8002} & \cellcolor{lightbluebar}\textbf{0.1369} & \cellcolor{lightbluebar}\textbf{264}s & \cellcolor{lightbluebar}\textbf{1.75}$\times$ & \cellcolor{lightbluebar}\textbf{799s} & \cellcolor{lightbluebar}\textbf{1.29$\times$} \\
& SVG2 & 77.90\% & 0.0830 & 21.80 & 0.7426 & 0.1707 & {411}s & {1.12}$\times$ & 883s & 1.17$\times$ \\
& \cellcolor{lightbluebar}SVG2+Ours & \cellcolor{lightbluebar}\textbf{78.22\%} & \cellcolor{lightbluebar}\textbf{0.0877} & \cellcolor{lightbluebar}\textbf{24.61} & \cellcolor{lightbluebar}\textbf{0.8178} & \cellcolor{lightbluebar}\textbf{0.1234} & \cellcolor{lightbluebar}\textbf{355}s & \cellcolor{lightbluebar}\textbf{1.30}$\times$ & \cellcolor{lightbluebar}\textbf{826s} & \cellcolor{lightbluebar}\textbf{1.25$\times$} \\
\midrule
\multirow{5}{*}{HunyuanVideo-13B} & Dense & 71.18\% & 0.0686 & -- & -- & -- & {385}s & {1.00}$\times$ & {630}s & {1.00}$\times$ \\
& XAttention & 71.55\% & 0.0440 & 26.06 & 0.8874 & 0.1348 & {249}s & {1.55}$\times$ & {497}s & {1.27}$\times$ \\
& \cellcolor{lightbluebar}XAttn+Ours & \cellcolor{lightbluebar}\textbf{71.89\%} & \cellcolor{lightbluebar}\textbf{0.0592} & \cellcolor{lightbluebar}\textbf{28.92} & \cellcolor{lightbluebar}\textbf{0.9181} & \cellcolor{lightbluebar}\textbf{0.0743} & \cellcolor{lightbluebar}\textbf{225}s & \cellcolor{lightbluebar}\textbf{1.71}$\times$ & \cellcolor{lightbluebar}\textbf{479}s & \cellcolor{lightbluebar}\textbf{1.32}$\times$ \\
& SVG2 & 69.89\% & 0.0530 & 30.21 & 0.9059 & 0.1237 & {305}s & {1.26}$\times$ & {539}s & {1.17}$\times$ \\
& \cellcolor{lightbluebar}SVG2+Ours & \cellcolor{lightbluebar}\textbf{70.02\%} & \cellcolor{lightbluebar}\textbf{0.0587} & \cellcolor{lightbluebar}\textbf{31.09} & \cellcolor{lightbluebar}\textbf{0.9125} & \cellcolor{lightbluebar}\textbf{0.1203} & \cellcolor{lightbluebar}\textbf{284}s & \cellcolor{lightbluebar}\textbf{1.36}$\times$ & \cellcolor{lightbluebar}\textbf{510}s & \cellcolor{lightbluebar}\textbf{1.24}$\times$ \\
\bottomrule
\end{tabular}
}
\caption{Main quantitative comparison among the original dense models, representative sparse attention methods, and the corresponding methods enhanced with our method. The first, third, and fourth blocks report 480P results, and the second block reports 720P results on Wan2.1-1.3B.}
\label{tab:main-results}
\end{table*}

\subsection{Error-guided Budgeted Calibration}
Observation~\ref{sec:obs2} shows that applying the same threshold to different heads can yield markedly different sparsity levels and sparse-induced velocity errors. EBC therefore calibrates thresholds at head granularity. For head $(l,h)$ and threshold $\tau$, let $s_{l,h}(\tau)$ and $e_{l,h}(\tau)$ denote the realized sparsity and induced velocity error. The exact joint velocity error under multi-head sparsification is generally non-separable, because perturbations from different heads can interact through downstream layers. To keep calibration tractable, we use a measurement-driven additive surrogate motivated by first-order additive propagation, and select head-specific thresholds under a global sparsity budget.

Our calibration procedure is designed with three considerations. First, the offline measurement cost should remain manageable, so we construct a discrete candidate set per head, cache the dense denoising velocities once, and measure each candidate with only one additional sparse forward while keeping all other heads dense. Second, calibration should reduce overfitting to a particular prompt or timestep, so we average measurements over sampled prompt-step pairs using both prompt sampling and interval-based timestep sampling. Third, the calibration objective should better reflect the actual effect of sparse attention on the generated videos, so we measure error on the denoising velocity rather than on the local attention output, and further use a frequency-aware weighted objective which is discussed later.

Concretely, we instantiate calibration on a discrete candidate set and solve a one-choice-per-head assignment problem. Each candidate \(k\) corresponds to a measured operating point, i.e., a threshold together with its resulting sparsity and velocity error. Let \(x_{l,h,k}\in\{0,1\}\) indicates whether head \((l,h)\) selects threshold candidate \(k\), then the offline selection can be written as
\begin{equation}
\begin{aligned}
\min_{\{x_{l,h,k}\}} \quad
& \sum_{l,h,k} E_{l,h,k}x_{l,h,k} \\
\text{s.t.}\quad
& \sum_k x_{l,h,k}=1,\ \forall l,h, \\
& \frac{1}{LH}\sum_{l,h,k} S_{l,h,k}x_{l,h,k}\ge S_{\min},
\end{aligned}
\end{equation}
where \(E_{l,h,k}\) and \(S_{l,h,k}\) are the measured velocity error and realized sparsity of candidate \(k\), and \(S_{\min}\) is the target global sparsity. This formulation is simple, but it is enough to express the actual design goal of EBC: keep the same global sparsity budget while redistributing thresholds across heterogeneous heads.

We measure calibration error on the denoising velocity rather than on the attention output, because downstream layers can attenuate or amplify local perturbations before they affect sampling. However, not all velocity errors have the same impact on final video quality. Inspired by \citep{liu2025freqca}, we therefore examine whether errors in different spatiotemporal frequency regions of the denoising velocity are equally harmful. We conduct a controlled perturbation study by injecting equal-relative-magnitude perturbations into different 3D-FFT regions and evaluating the resulting video degradation against the dense reference. As shown in Table~\ref{tab:frequency-sensitivity}, perturbations in the low-temporal low-spatial region are substantially more harmful than those in higher-frequency regions. This suggests that treating all velocity-space errors uniformly is suboptimal. We therefore use a weighted four-band 3D-FFT objective in the main experiments. Specifically, for head $(l,h)$ under threshold $\tau$, let $y^{\mathrm{sparse}}_{l,h}(\tau)$ denote the denoising velocity obtained when head $(l,h)$ uses threshold $\tau$ and all other heads remain dense, and let $y^{\mathrm{dense}}$ denote the dense reference velocity. We define the sparse-induced velocity error as
\begin{equation}
\epsilon_{l,h}(\tau)
:=
y^{\mathrm{sparse}}_{l,h}(\tau)-y^{\mathrm{dense}}.
\end{equation}
Let \(\widehat{\epsilon}_{l,h}(\tau,\omega)\) and \(\widehat{y}^{\mathrm{dense}}(\omega)\) denote the values of the 3D discrete Fourier transforms of \(\epsilon_{l,h}(\tau)\) and \(y^{\mathrm{dense}}\), respectively, at frequency index \(\omega\). We partition the frequency grid into four bands: low-temporal low-spatial (LL), low-temporal high-spatial (LH), high-temporal low-spatial (HL), and high-temporal high-spatial (HH), denoted by \(\{\Omega_q\}_{q=1}^4\). The calibration error is then defined as
\begin{equation}
e_{l,h}(\tau)
:=
\sum_{q=1}^{4} w_q \,
\frac{\sum_{\omega\in\Omega_q}\left|\widehat{\epsilon}_{l,h}(\tau,\omega)\right|^2}
{\sum_{\omega}\left|\widehat{y}^{\mathrm{dense}}(\omega)\right|^2+\varepsilon},
\end{equation}
where \(\left|\widehat{\epsilon}_{l,h}(\tau,\omega)\right|^2\) is the spectral energy of the sparse-induced error at frequency \(\omega\), \(w_q\) is the weight assigned to band \(\Omega_q\), and \(\varepsilon\) is a small constant for numerical stability.

\section{Experiments}

\subsection{Experimental Setup}

\noindent\textbf{Baselines.} We instantiate the method on XAttention \citep{xu2025xattention} and SVG2 \citep{yang2026sparse}. Following prior sparse-attention methods such as XAttention, SVG, and SVG2 \citep{xu2025xattention,xi2025sparse,yang2026sparse}, all experiments keep the first five denoising steps in dense attention. This early dense-attention warm-up is crucial for preserving similarity to the original dense-generation outputs.

\noindent\textbf{Model setting.} We evaluate Wan2.1-1.3B \citep{wan2025wan} with 81 frames at both \(832\times480\) and \(1280\times720\) resolutions, Wan2.1-14B \citep{wan2025wan} with 81 frames at \(832\times480\), and HunyuanVideo-13B \citep{kong2024hunyuanvideo} with 129 frames at \(720\times480\). All experiments use 50 denoising steps on NVIDIA A800 GPUs.

\noindent\textbf{Evaluation metrics.} We report VBench \citep{huang2024vbench} and VisionReward \citep{xu2026visionreward} for perceptual quality, PSNR/SSIM/LPIPS \citep{wang2004image,zhang2018unreasonable} for similarity to dense outputs, and latency/speedup for efficiency. We report both attention-only and end-to-end efficiency. Here, Attn Latency includes all attention-related operations, including sparse-mask prediction, mask reuse/refresh decisions, and sparse attention computation, whereas E2E Latency denotes the end-to-end runtime of the DiT denoising model. We note that, for SVG2, the end-to-end speedup is not attributable solely to attention acceleration, but also reflects the optimized RoPE and normalization implementations provided by the original SVG2 framework. The reported VBench score follows the official weighting protocol over subject consistency, motion smoothness, dynamic degree, aesthetic quality, imaging quality, and overall consistency.

\subsection{Main Results}

We compare HEART with dense attention and representative training-free sparse attention methods, including XAttention \citep{xu2025xattention} and SVG2 \citep{yang2026sparse}. At 480P on Wan2.1-1.3B, augmenting XAttention raises attention-only and DiT denoising speedup from $1.55\times$ and $1.27\times$ to $2.23\times$ and $1.44\times$, respectively, while improving VBench, VisionReward, and all three similarity metrics. HEART similarly improves SVG2 from $1.22\times$/$1.25\times$ to $1.49\times$/$1.39\times$ in attention-only/DiT speedup, together with a PSNR gain from 23.84 to 24.78.

The benefit remains clear at higher resolution and larger model scales. At 720P, HEART increases XAttention's attention-only speedup from $2.23\times$ to $2.68\times$ and DiT denoising speedup from $1.67\times$ to $1.85\times$, while improving every reported quality and similarity metric. On HunyuanVideo-13B, it raises XAttention's PSNR from 26.06 to 28.92 and its DiT denoising speedup from $1.27\times$ to $1.32\times$. These consistent gains show that TMR and EBC improve the quality--efficiency frontier rather than merely exchanging fidelity for runtime.

\subsection{Ablation Studies}

All ablation experiments are conducted on Wan2.1-1.3B at 480P using XAttention~\cite{xu2025xattention} under the same inference configuration as the main experiment.

\subsubsection{Key Components Ablation}

We isolate the contribution of the two proposed modules. Table~\ref{tab:component-ablation} shows that TMR and EBC play clearly different but complementary roles. TMR primarily improves efficiency by removing redundant online mask updates, which raises the speedup substantially. EBC, in contrast, mainly improves quality under nearly the same computational budget, indicating that better head-wise threshold allocation can recover fidelity without relying on additional runtime. Combining the two yields the best overall trade-off.

\begin{table}[t]
\centering
{\small
\begin{tabular*}{\columnwidth}{@{\extracolsep{\fill}}lcccc@{}}
\toprule
\multirow{2}{*}{Variant} & \multicolumn{3}{c}{Similarity} & Efficiency \\
\cmidrule(lr){2-4}\cmidrule(lr){5-5}
& PSNR$\uparrow$ & SSIM$\uparrow$ & LPIPS$\downarrow$ & E2E Spd.$\uparrow$ \\
\midrule
XAttention & 21.04 & 0.6481 & 0.2527 & 1.27$\times$ \\
+ TMR only & 21.46 & 0.6774 & 0.2355 & 1.43$\times$ \\
+ EBC only & 21.80 & 0.6846 & 0.2310 & 1.28$\times$ \\
+ HEART & 21.61 & 0.6833 & 0.2337 & 1.44$\times$ \\
\bottomrule
\end{tabular*}
}
\caption{Component ablation of HEART on Wan2.1-1.3B at 480P with XAttention, isolating the complementary efficiency benefit of TMR and fidelity benefit of EBC.}
\label{tab:component-ablation}
\end{table}

\begin{table}[t]
\centering
{\small
\begin{tabular*}{\columnwidth}{@{\extracolsep{\fill}}lcccc@{}}
\toprule
\multicolumn{5}{c}{\textbf{Reuse Policy Ablation}} \\
\midrule
\multirow{2}{*}{Reuse policy} & \multicolumn{3}{c}{Similarity} & Efficiency \\
\cmidrule(lr){2-4}\cmidrule(lr){5-5}
& PSNR$\uparrow$ & SSIM$\uparrow$ & LPIPS$\downarrow$ & E2E Spd.$\uparrow$ \\
\midrule
Interval Reuse & 21.19 & 0.6774 & 0.2380 & 1.50$\times$ \\
Warmup Reuse & 21.38 & 0.6744 & 0.2365 & 1.44$\times$ \\
TMR (Ours) & 21.61 & 0.6833 & 0.2337 & 1.44$\times$ \\
\midrule
\multicolumn{5}{c}{\textbf{TMR Threshold Ablation}} \\
\midrule
\multirow{2}{*}{$\delta$} & \multicolumn{3}{c}{Similarity} & Efficiency \\
\cmidrule(lr){2-4}\cmidrule(lr){5-5}
& PSNR$\uparrow$ & SSIM$\uparrow$ & LPIPS$\downarrow$ & E2E Spd.$\uparrow$ \\
\midrule
0.0 & 21.80 & 0.6846 & 0.2310 & 1.28$\times$ \\
10.0 & 21.57 & 0.6825 & 0.2347 & 1.39$\times$ \\
30.0 & 21.61 & 0.6833 & 0.2337 & 1.44$\times$ \\
100.0 & 20.49 & 0.6659 & 0.2702 & 1.34$\times$ \\
\midrule
\multicolumn{5}{c}{\textbf{Calibration Objective Ablation}} \\
\midrule
\multirow{2}{*}{Objective} & \multicolumn{3}{c}{Similarity} & Efficiency \\
\cmidrule(lr){2-4}\cmidrule(lr){5-5}
& PSNR$\uparrow$ & SSIM$\uparrow$ & LPIPS$\downarrow$ & E2E Spd.$\uparrow$ \\
\midrule
Velocity MSE & 21.54 & 0.6773 & 0.2369 & 1.44$\times$ \\
Weighted 3D-FFT & 21.61 & 0.6833 & 0.2337 & 1.44$\times$ \\
\bottomrule
\end{tabular*}
}
\caption{Comprehensive ablations of TMR and EBC on Wan2.1-1.3B at 480P with XAttention. Top: comparison of our head-wise adaptive TMR with fixed-interval and warmup reuse policies. Middle: effect of the query-key drift threshold $\delta$, which determines when cached masks are refreshed. Bottom: comparison of the velocity-MSE and frequency-weighted 3D-FFT calibration objectives under the same global sparsity budget.}
\label{tab:tmr-objective-ablation}
\end{table}

\subsubsection{Ablation on the Reuse Strategy}

We compare TMR with two shared reuse schedules: fixed-interval refresh from AdaSpa \citep{xia2025training} and warmup reuse from PAROAttention \citep{zhao2026paroattention}. As shown in Table~\ref{tab:tmr-objective-ablation}, Interval Reuse attains higher speedup, but sacrifices similarity. Moreover, AdaSpa reports that even reducing the refresh interval does not recover better quality; we conjecture that this is because a fixed interval-based reuse rule is overly heuristic and therefore cannot adapt well to the actual evolution of sparse masks \citep{xia2025training}. Warmup Reuse is also weaker in the speed--quality trade-off: even when matched at a similar speedup, it still produces worse similarity metrics than TMR. This suggests that shared reuse schedules are inherently too heuristic, whereas effective mask reuse should adapt to the head-wise and step-wise variation of mask evolution.

\subsubsection{Ablation on the Reuse Threshold}

The reuse threshold $\delta$ controls the aggressiveness of TMR and induces a clear trade-off. Moderate reuse gives the best balance: compared with no reuse, \(\delta=30.0\) substantially improves speedup while keeping the similarity degradation small. However, pushing reuse too aggressively is harmful. When \(\delta\) becomes too large, stale masks are reused beyond their valid range, which weakens both fidelity and the final speedup. This behavior is consistent with the role of TMR: reuse is beneficial only when it remains adaptive to the evolving sparse pattern.

\subsubsection{Ablation on the Calibration Objective}

We compare raw velocity-space MSE with the weighted 3D-FFT objective used in the main method under the same sparsity budget. The weighted spectral objective consistently improves all three similarity metrics, suggesting that emphasizing low-frequency perturbations leads to better structural fidelity in the final output. This supports our motivation for using a frequency-aware calibration target rather than a plain pointwise velocity error.

\section{Conclusion}

In this paper, we revisit two practical bottlenecks of training-free sparse attention for video diffusion: frequent sparse-mask reconstruction and the suboptimal use of a shared threshold across heterogeneous heads. HEART addresses them with two complementary components: Temporal Mask Reuse, which adaptively reuses sparse masks based on a lightweight query-key drift signal, and Error-guided Budgeted Calibration, which assigns head-specific thresholds under a global sparsity budget using a frequency-aware velocity-error objective. As plug-in modifications to existing sparse-attention pipelines, these two components consistently improve the quality--efficiency frontier across sparse-attention backbones, model scales, and resolutions, without retraining or kernel changes.

\clearpage
\bibliographystyle{plainnat}
\bibliography{references}

\begin{figure}[H]
\centering
\includegraphics[width=0.48\textwidth]{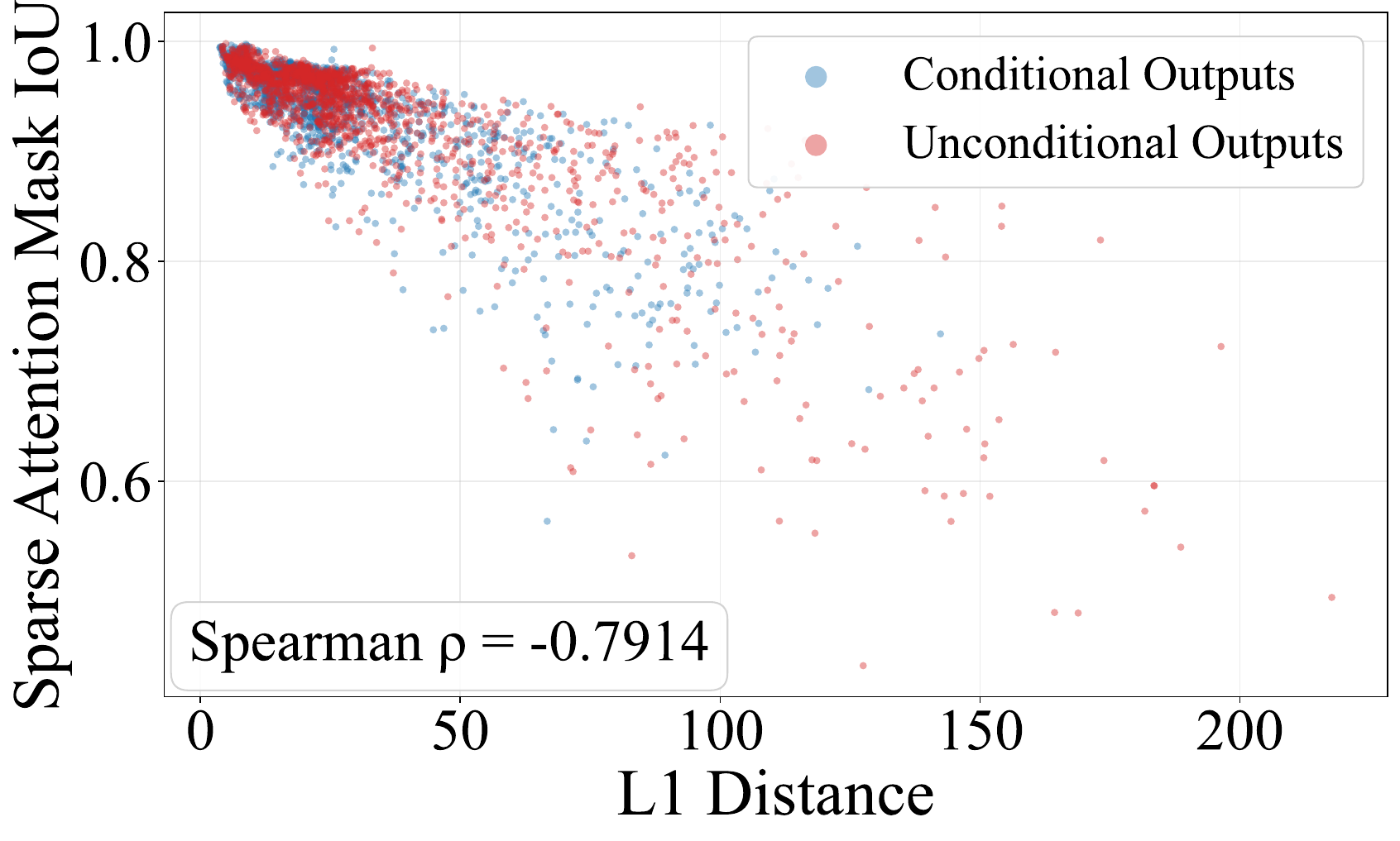}\hfill
\includegraphics[width=0.48\textwidth]{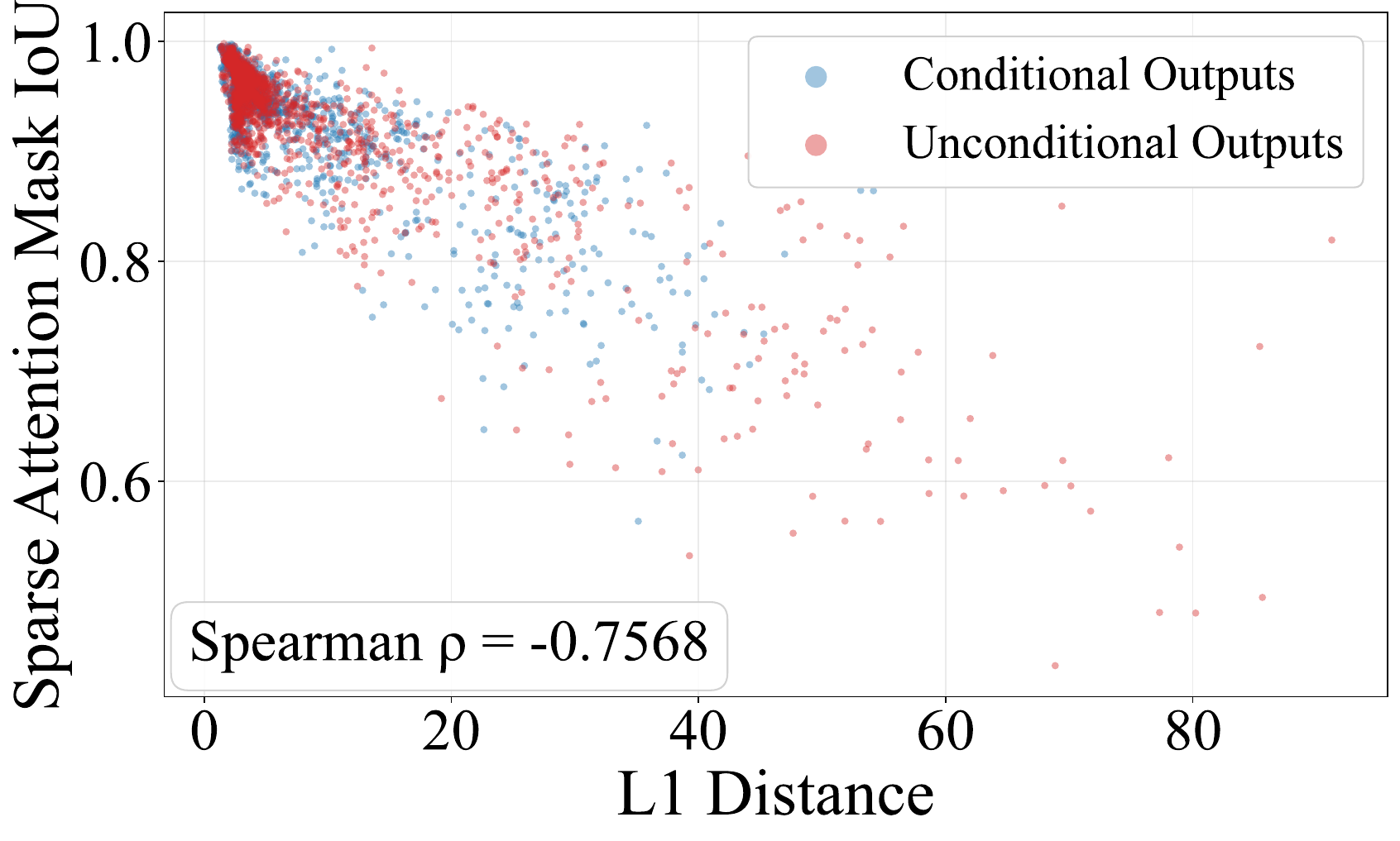}
\caption{Relationship between adjacent-step mask IoU and query-key drift. The full-token drift and the mean-pooled proxy both show strong negative correlation with mask IoU, indicating that the cheaper statistic preserves the predictive trend needed by TMR while requiring much less cache memory.}
\label{fig:iou-vs-qk-drift}
\end{figure}

\clearpage
\beginappendix
\long\def\TheoreticalDetails{
\section{Theoretical Discussion and Bounds}
\subsection{Stability of Query-Key Interactions}
\label{app:tmr_theory}

For an attention head \(h\), let \(Q_t^{(h)},K_t^{(h)}\in\mathbb{R}^{N\times D}\) be the query and key matrices at denoising step \(t\). Here \(N\) is the number of tokens and \(D\) is the head dimension. Let \(q_{t,i}^{(h)},k_{t,j}^{(h)}\in\mathbb{R}^{D}\) denote the \(i\)-th query row and \(j\)-th key row, respectively. We compare an anchor step \(t_a\) with a later step \(t_b>t_a\).

\paragraph{Query-key interaction.}
The unnormalized query-key interaction matrix is
\begin{equation}
G_t^{(h)}
 := \frac{Q_t^{(h)}K_t^{(h)\top}}{\sqrt{D}}
 \in\mathbb{R}^{N\times N},
\end{equation}
and its \((i,j)\)-th entry is
\begin{equation}
g_t^{(h)}(i,j)
 := \frac{q_{t,i}^{(h)\top}k_{t,j}^{(h)}}{\sqrt{D}}.
\end{equation}
Thus, \(G_t^{(h)}\) contains all pairwise query-key interactions for head \(h\) at step \(t\).

\paragraph{Pointwise stability bound.}
Define the feature changes
\[
\Delta q_i^{(h)} := q_{t_b,i}^{(h)}-q_{t_a,i}^{(h)},
\qquad
\Delta k_j^{(h)} := k_{t_b,j}^{(h)}-k_{t_a,j}^{(h)}.
\]
For a fixed pair \((i,j)\), the interaction difference can be expanded as
\begin{align}
g_{t_b}^{(h)}(i,j)-g_{t_a}^{(h)}(i,j)
 &= \frac{
\Delta q_i^{(h)\top}k_{t_a,j}^{(h)}
 +q_{t_b,i}^{(h)\top}\Delta k_j^{(h)}
}{\sqrt{D}}.
\end{align}
By Hölder's inequality,
\begin{align}
\left|g_{t_b}^{(h)}(i,j)-g_{t_a}^{(h)}(i,j)\right|
 &\leq \frac{1}{\sqrt{D}}
\left\|k_{t_a,j}^{(h)}\right\|_\infty
\left\|\Delta q_i^{(h)}\right\|_1
\notag\\
 &\quad+
\frac{1}{\sqrt{D}}
\left\|q_{t_b,i}^{(h)}\right\|_\infty
\left\|\Delta k_j^{(h)}\right\|_1.
\label{eq:app_qk_pointwise_bound}
\end{align}
Assume that the query and key features are bounded over the two steps:
\[
\left\|q_{t,i}^{(h)}\right\|_\infty\leq B_q,
\qquad
\left\|k_{t,j}^{(h)}\right\|_\infty\leq B_k,
\qquad
t\in\{t_a,t_b\}.
\]
Then
\begin{equation}
\left|g_{t_b}^{(h)}(i,j)-g_{t_a}^{(h)}(i,j)\right|
 \leq \frac{
B_k\left\|\Delta q_i^{(h)}\right\|_1
 +B_q\left\|\Delta k_j^{(h)}\right\|_1
}{\sqrt{D}}.
\label{eq:app_qk_bounded_bound}
\end{equation}

\paragraph{Averaged interaction drift.}
Define the full-token query-key drift
\begin{equation}
d_{t_a\rightarrow t_b}^{(h)}
 := \frac{1}{N}\sum_{i=1}^{N}\left\|\Delta q_i^{(h)}\right\|_1
 +\frac{1}{N}\sum_{j=1}^{N}\left\|\Delta k_j^{(h)}\right\|_1.
\label{eq:app_full_token_drift}
\end{equation}
Averaging Eq.~\eqref{eq:app_qk_bounded_bound} over all \(N^2\) query-key pairs gives
\begin{align}
\Delta_{\mathrm{QK}}^{(h)}
 &:= \frac{1}{N^2}\sum_{i=1}^{N}\sum_{j=1}^{N}
\left|g_{t_b}^{(h)}(i,j)-g_{t_a}^{(h)}(i,j)\right| \notag\\
 &\leq \frac{B_k}{\sqrt{D}}\frac{1}{N}\sum_{i=1}^{N}
\left\|\Delta q_i^{(h)}\right\|_1 \notag\\
 &\quad+\frac{B_q}{\sqrt{D}}\frac{1}{N}\sum_{j=1}^{N}
\left\|\Delta k_j^{(h)}\right\|_1 \notag\\
 &\leq \frac{\max\{B_q,B_k\}}{\sqrt{D}}\,
d_{t_a\rightarrow t_b}^{(h)}.
\label{eq:app_qk_average_bound}
\end{align}
Therefore, the average change in the full query-key interaction matrix is bounded by the average changes in the query and key features. This is the theoretical justification for using their drift as a stability signal.

\paragraph{Relation to the online proxy.}
The full-token drift in Eq.~\eqref{eq:app_full_token_drift} is useful for analysis but expensive to cache. TMR therefore uses the mean-pooled proxy
\begin{equation}
\widetilde d_{t_a\rightarrow t_b}^{(h)}
 := \left\|\bar Q_{t_b}^{(h)}-\bar Q_{t_a}^{(h)}\right\|_1
 +\left\|\bar K_{t_b}^{(h)}-\bar K_{t_a}^{(h)}\right\|_1,
\label{eq:app_mean_pool_drift}
\end{equation}
where \(\bar Q_t^{(h)},\bar K_t^{(h)}\in\mathbb{R}^{D}\) are the token means of \(Q_t^{(h)}\) and \(K_t^{(h)}\), respectively. The derivation above explains why query-key drift is a meaningful stability signal. Blocks near the top-\(p\) boundary are marginal by construction: they are included mainly to complete the required cumulative importance mass, so their individual contributions are typically smaller than those of the core retained blocks. A small interaction perturbation may therefore exchange a few boundary blocks and reduce exact mask IoU while changing the total retained importance and attention output only slightly. This makes boundary-level changes less consequential than changes involving high-importance blocks. The negative empirical correlation with adjacent-step mask IoU in Figure~\ref{fig:iou-vs-qk-drift} further supports using the cheaper mean-pooled version in practice.

\subsection{Additive Calibration Objective as a Local Surrogate}
\label{sec:appendix-additive-surrogate}

The EBC objective sums velocity errors measured by sparsifying one head at a time. This is not an exact decomposition of the error obtained when many heads are sparsified jointly. The following derivation explains why the resulting additive objective is nevertheless a useful local surrogate: head perturbations are exactly additive at the attention output and remain additive to first order after propagation to the denoising velocity, while the scalar velocity error introduces cross-head interaction terms.

\paragraph{Head-wise perturbations.}
Consider layer $l$ with $H$ attention heads. Let $o_{l,h}$ be the output of head $h$ before the output projection, and let
\begin{equation}
A_l = \mathrm{Concat}(o_{l,1}, \dots, o_{l,H}) W_O
\end{equation}
be the multi-head attention output after the projection matrix $W_O$. If sparse attention induces a perturbation $\delta_{l,h}$ on head $h$, then
\begin{equation}
A_l^{\mathrm{sparse}}
=
\mathrm{Concat}(o_{l,1}+\delta_{l,1}, \dots, o_{l,H}+\delta_{l,H}) W_O.
\end{equation}
Writing $W_{O,h}$ for the block of $W_O$ associated with head $h$, the attention-output perturbation is
\begin{equation}
\begin{aligned}
\Delta A_l
&:=
A_l^{\mathrm{sparse}} - A_l \\
&=
\mathrm{Concat}(\delta_{l,1}, \dots, \delta_{l,H}) W_O \\
&=
\sum_{h=1}^H \delta_{l,h} W_{O,h},
\end{aligned}
\label{eq:appendix-attn-perturb-additive}
\end{equation}
Define $\Delta A_{l,h}:=\delta_{l,h}W_{O,h}$, so that
\begin{equation}
\Delta A_l = \sum_{h=1}^H \Delta A_{l,h}.
\end{equation}

\paragraph{Propagation to denoising velocity.}
Let $\mathcal{F}_l$ map the attention output $A_l$ of layer $l$ to the denoising velocity $y$, with all other model computations fixed. Around the dense reference point, assume that $\mathcal{F}_l$ is twice continuously differentiable. For sufficiently small $\Delta A_l$,
\begin{equation}
\begin{aligned}
\Delta y_l
&:= y^{\mathrm{sparse}} - y^{\mathrm{dense}} \\
&= \mathcal{F}_l(A_l + \Delta A_l) - \mathcal{F}_l(A_l) \\
&= J_l \Delta A_l + R_l,
\end{aligned}
\label{eq:appendix-output-taylor}
\end{equation}
where $J_l$ is the Jacobian of $\mathcal{F}_l$ at $A_l$ and $R_l$ is the higher-order remainder. If the Hessian of $\mathcal{F}_l$ is bounded near $A_l$, then
If the Hessian of $\mathcal{F}_l$ is bounded in a neighborhood of $A_l$, then there exists a constant $\beta_l$
such that
\begin{equation}
\|R_l\|
\le
\frac{\beta_l}{2}\|\Delta A_l\|^2.
\label{eq:appendix-output-remainder}
\end{equation}
Thus, with $u_{l,h}:=J_l\Delta A_{l,h}$,
\begin{equation}
\Delta y_l = \sum_{h=1}^H u_{l,h} + R_l.
\label{eq:appendix-vector-additive}
\end{equation}
The propagated velocity perturbation is therefore additive to first order, up to $R_l$.

\paragraph{Why scalar velocity error is not additive.}
EBC uses a scalar error on the denoising velocity, such as velocity-space MSE or the weighted spectral error. Let $\psi$ map the velocity perturbation to this nonnegative scalar error, with $\psi(0)=0$:
\begin{equation}
e_l = \psi(\Delta y_l).
\end{equation}
For the squared-error objectives used here, the leading term around $\Delta y_l=0$ is quadratic. If $\psi$ is twice continuously differentiable, then
\begin{equation}
e_l = \frac{1}{2}\Delta y_l^\top H_\psi(\xi)\Delta y_l,
\label{eq:appendix-scalar-taylor}
\end{equation}
where $H_\psi(\xi)$ is the Hessian at an intermediate point. Ignoring $R_l$ temporarily and substituting Eq.~\eqref{eq:appendix-vector-additive} gives
\begin{equation}
e_l
\approx
\frac{1}{2}
\left(
\sum_{h=1}^H u_{l,h}
\right)^\top
H_\psi(\xi)
\left(
\sum_{h=1}^H u_{l,h}
\right).
\end{equation}
Expanding this expression gives
\begin{equation}
e_l
\approx
\frac{1}{2}\sum_{h=1}^H u_{l,h}^\top H_\psi(\xi) u_{l,h}
+
\frac{1}{2}\sum_{\substack{h,h'=1\\h\neq h'}}^H
u_{l,h}^\top H_\psi(\xi) u_{l,h'}.
\label{eq:appendix-scalar-decomposition}
\end{equation}
The first sum contains head-wise self terms, whereas the second contains cross-head interaction terms. Hence the joint scalar velocity error is generally not additive, even though the propagated velocity perturbation is additive to first order.

\paragraph{Measurement-driven additive surrogate.}
A local additive surrogate keeps the head-wise self terms and drops the interaction terms:
\begin{equation}
e_l^{\mathrm{diag}}
:=
\sum_{h=1}^H \widetilde e_{l,h},
\qquad
\widetilde e_{l,h}
:=
\frac{1}{2}u_{l,h}^\top H_\psi(\xi_h) u_{l,h},
\label{eq:appendix-diag-surrogate}
\end{equation}
where $\xi_h$ is a local point associated with head $h$. This is a diagonalized local approximation, not an exact decomposition of the full-network error.

In practice, we do not estimate $\widetilde e_{l,h}$ from local curvature. For each candidate threshold, we instead measure the isolated single-head velocity error by sparsifying only head $(l,h)$ while keeping all other heads dense:
\begin{equation}
E_{l,h,k}
:=
\psi\!\left(y_{l,h,k}^{\mathrm{sparse}}-y^{\mathrm{dense}}\right),
\end{equation}
where $k$ indexes the candidate threshold. The offline ILP optimizes the resulting measurement-driven surrogate
\begin{equation}
\sum_{l=1}^L \sum_{h=1}^H \sum_{k=1}^K E_{l,h,k} x_{l,h,k},
\end{equation}
which sums isolated single-head velocity errors. The approximation ignores cross-head interactions, higher-order propagation through $R_l$, and cross-layer coupling. It is therefore best understood as a tractable local surrogate that preserves head-wise velocity sensitivity without claiming an exact factorization of the full-network error.
}

\section{Method Details}
\subsection{Further Evidence for Observation I: Head-Wise Heterogeneity in Mask Evolution}

To complement the head-level evidence in the main paper, we further examine temporal mask similarity at coarser granularities. Specifically, we analyze prompt-level IoU curves obtained by averaging over all layers and heads, and layer-level heatmaps for a randomly selected prompt. These views help verify whether the temporal heterogeneity observed in the main paper is confined to a few heads or persists more broadly across prompts and layers.

\begin{figure}[H]
\centering
\includegraphics[width=0.82\textwidth]{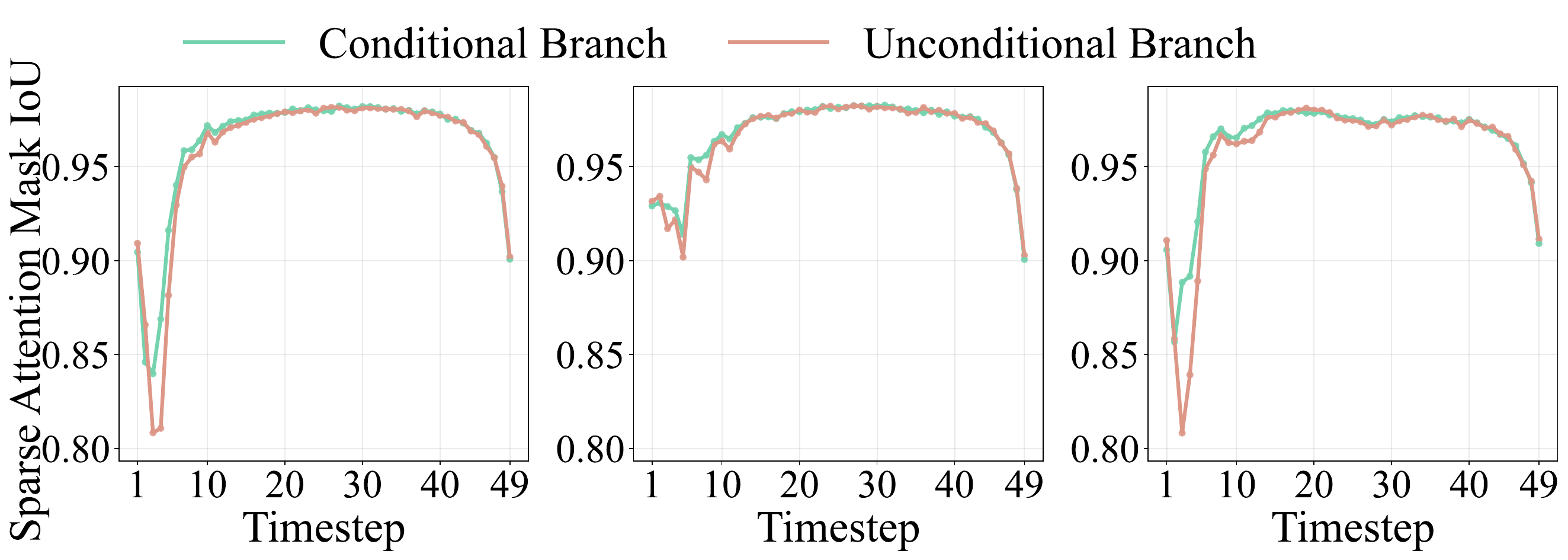}\\[0.4em]
{\small \textbf{(a)} Prompt-level IoU curves, obtained by averaging over all layers and heads, still vary substantially across prompts.}\\[0.8em]
\includegraphics[width=0.72\textwidth]{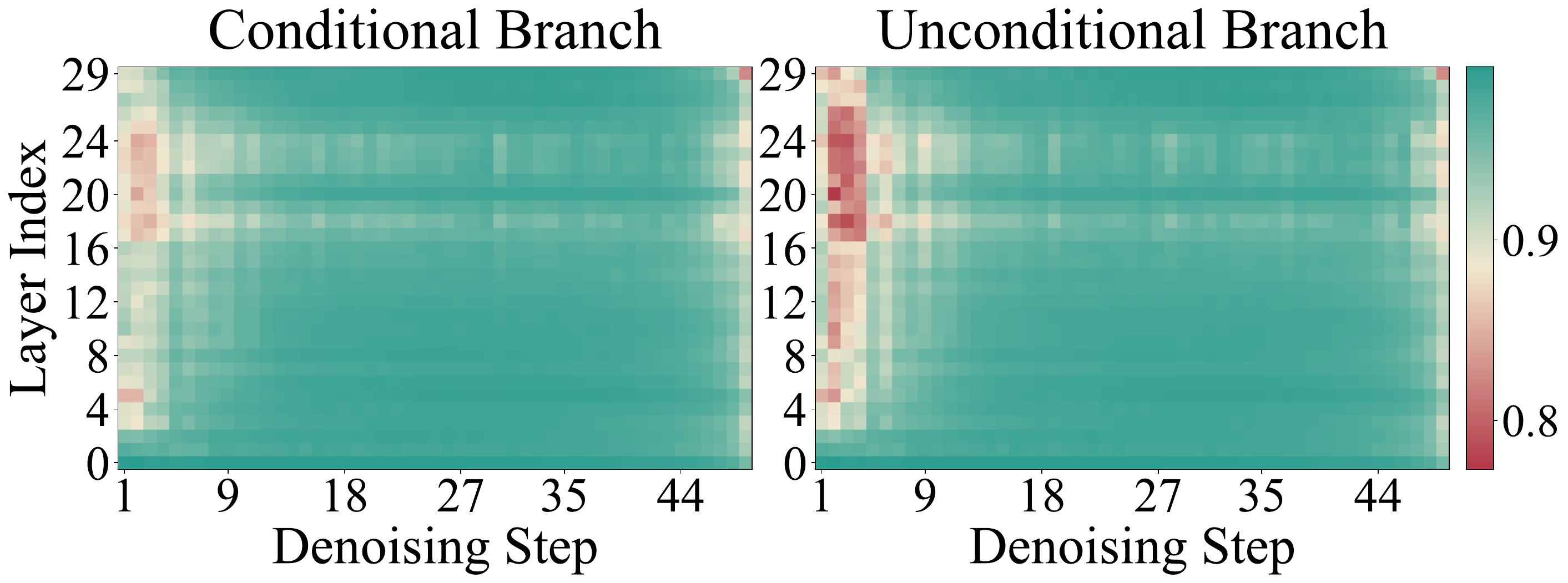}\\[0.4em]
{\small \textbf{(b)} Layer-step heatmap for a randomly selected prompt shows that the heterogeneity also persists across layers.}
\caption{Temporal mask-similarity views supporting Observation I. Together with the head-level view in the main paper, these results support adaptive head-wise mask reuse.}
\label{fig:temporal-mask-similarity-supp}
\end{figure}

As shown in Figure~\ref{fig:temporal-mask-similarity-supp}, the heterogeneity indeed persists beyond individual heads. At the prompt level, different prompts exhibit noticeably different adjacent-step mask-IoU trajectories. At the layer level, some layers remain relatively stable over multiple denoising steps, whereas others vary much more rapidly. Together with the head-level visualization in the main paper, these results support adaptive reuse rather than predefined shared schedules.

\subsection{Why Error-guided Budgeted Calibration Measures Multiple Timesteps}

\begin{figure}[H]
\centering
\includegraphics[width=\columnwidth]{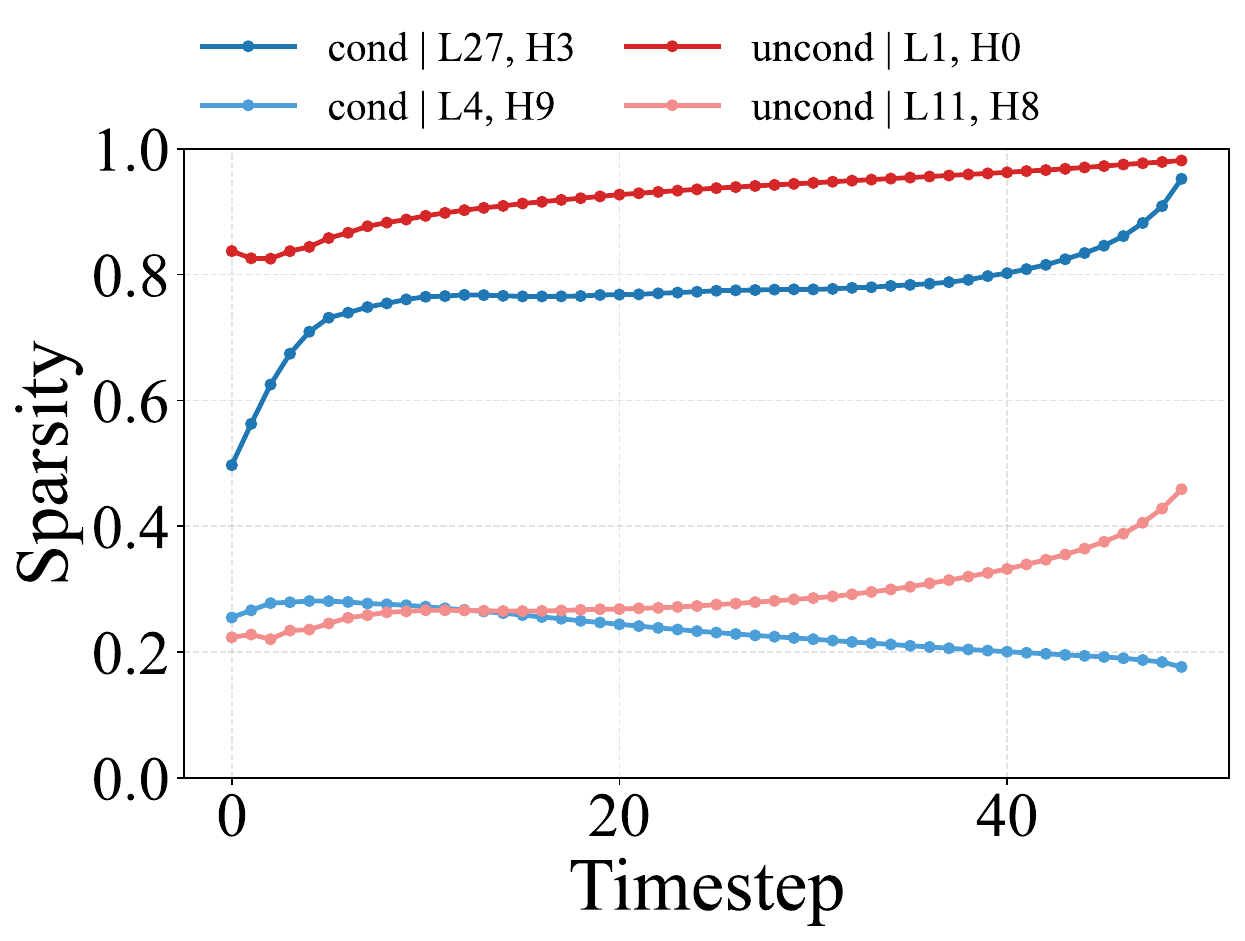}
\caption{Realized sparsity over denoising timesteps for several randomly selected heads under a fixed sparsification threshold. Even for the same head, the achieved sparsity varies substantially across timesteps, while different heads exhibit different trajectories. This motivates interval-based timestep sampling in calibration, rather than measuring each head at only one denoising step.}
\label{fig:head-sparsity-over-timesteps}
\end{figure}

Figure~\ref{fig:head-sparsity-over-timesteps} shows that realized sparsity under a fixed threshold can vary substantially even for the same head across denoising timesteps. This is why the offline calibration stage samples multiple timesteps from the denoising trajectory, rather than relying on a single-step measurement.

\subsection{Temporal Mask Reuse Procedure}

Algorithm~\ref{alg:tmr} details the online head-wise mask reuse procedure. The layer-level gating heuristic described in the main paper is applied after these per-head decisions are obtained.

\long\def\TMRAlgorithm{
\begin{algorithm}[H]
\caption{Temporal Mask Reuse}
\label{alg:tmr}
\small
\begin{algorithmic}[1]
\REQUIRE Current $Q_{t_b}^{(h)}, K_{t_b}^{(h)}$, reuse threshold $\delta$, cached anchor features $\bar{Q}_{t_a}^{(h)}, \bar{K}_{t_a}^{(h)}$, cached anchor mask $M_{t_a}^{(h)}$
\STATE $\bar{Q}_{t_b}^{(h)} \gets \operatorname{mean}_N(Q_{t_b}^{(h)})$, \quad $\bar{K}_{t_b}^{(h)} \gets \operatorname{mean}_N(K_{t_b}^{(h)})$
\IF{no cached anchor exists for head $h$}
    \STATE $M_{t_b}^{(h)} \gets \textsc{PredictMask}(h,t_b)$
    \STATE Set anchor $t_a \gets t_b$
    \STATE Cache $(\bar{Q}_{t_a}^{(h)}, \bar{K}_{t_a}^{(h)}, M_{t_a}^{(h)}) \gets (\bar{Q}_{t_b}^{(h)}, \bar{K}_{t_b}^{(h)}, M_{t_b}^{(h)})$
\ELSE
    \STATE $d_{t_a\rightarrow t_b}^{(h)} \gets
    \|\bar{Q}_{t_a}^{(h)}-\bar{Q}_{t_b}^{(h)}\|_1
    +
    \|\bar{K}_{t_a}^{(h)}-\bar{K}_{t_b}^{(h)}\|_1$
    \IF{$d_{t_a\rightarrow t_b}^{(h)} > \delta$}
        \STATE $M_{t_b}^{(h)} \gets \textsc{PredictMask}(h,t_b)$
        \STATE Set anchor $t_a \gets t_b$
        \STATE Cache $(\bar{Q}_{t_a}^{(h)}, \bar{K}_{t_a}^{(h)}, M_{t_a}^{(h)}) \gets (\bar{Q}_{t_b}^{(h)}, \bar{K}_{t_b}^{(h)}, M_{t_b}^{(h)})$
    \ELSE
        \STATE $M_{t_b}^{(h)} \gets M_{t_a}^{(h)}$
        \STATE Keep the anchor cache unchanged
    \ENDIF
\ENDIF
\RETURN $M_{t_b}^{(h)}$
\end{algorithmic}
\end{algorithm}
}
\TMRAlgorithm
\FloatBarrier

\subsection{Error-guided Budgeted Calibration Procedure}

Algorithm~\ref{alg:head-calibration} details the offline measurement and calibration procedure used to obtain one top-$p$ threshold for each head.

\long\def\EBCAlgorithm{
\begin{algorithm}[H]
\caption{Error-guided Budgeted Calibration}
\label{alg:head-calibration}
\small
\begin{algorithmic}[1]
\REQUIRE Prompt pool $\mathcal{P}$, timestep intervals $\{\mathcal{I}_1,\dots,\mathcal{I}_J\}$, threshold candidates $\{\tau_1,\dots,\tau_K\}$, target sparsity $S_{\min}$
\FORALL{sampled prompts $p \in \mathcal{P}$}
    \STATE run dense inference once and cache dense denoising outputs $\{y^{\mathrm{dense}}(p,t)\}_{t}$
\ENDFOR
\FOR{$l = 1$ to $L$}
    \FOR{$h = 1$ to $H$}
        \STATE assign one prompt $p_{l,h}$ to head $(l,h)$
        \STATE sample timesteps $\{t_j\}_{j=1}^{J}$ with $t_j \sim \mathcal{I}_j$
        \FOR{$k = 1$ to $K$}
            \FOR{$j = 1$ to $J$}
                \STATE sparsify only head $(l,h)$ at $(p_{l,h}, t_j)$ with threshold $\tau_k$
                \STATE compute model-output error $e^{(j)}_{l,h,k}$ from cached dense output and sparse output
                \STATE compute realized sparsity $s^{(j)}_{l,h,k}$
            \ENDFOR
            \STATE $E_{l,h,k} \gets \frac{1}{J}\sum_{j=1}^{J} e^{(j)}_{l,h,k}$
            \STATE $S_{l,h,k} \gets \frac{1}{J}\sum_{j=1}^{J} s^{(j)}_{l,h,k}$
        \ENDFOR
    \ENDFOR
\ENDFOR
\STATE solve the ILP defined by $\{E_{l,h,k}, S_{l,h,k}\}$ under sparsity budget $S_{\min}$
\RETURN calibrated thresholds $\{\tau^\star_{l,h}\}$
\end{algorithmic}
\end{algorithm}
}

\EBCAlgorithm
\FloatBarrier

\newpage
\section{Experimental Details}

\subsection{XAttention and SVG2 Configurations}

We retain the original sparse-attention configurations of XAttention \citep{xu2025xattention} and SVG2 \citep{yang2026sparse}. For XAttention, we use an anti-diagonal scoring stride of 16 and a block size of 128 tokens. Blocks are selected by the original inverse top-$p$ rule with a shared threshold of $p=0.9$. For SVG2, we use $C_q=300$ query centroids and $C_k=1000$ key centroids, a shared top-$p$ threshold of $p=0.9$, a minimum retained-key-cluster ratio of 0.10, and 50 and 2 K-means iterations for initialization and subsequent steps, respectively. We use the optimized RoPE--Triton--normalization path provided by SVG2. Following XAttention, SVG, and SVG2, the first five denoising steps use dense attention, which is important for preserving similarity to the dense reference. No early transformer layers are kept dense, and HEART is applied without changing any other baseline-specific setting.

\subsection{HEART Configurations}
\label{sec:appendix-reproducibility}

\paragraph{TMR.}
Unless explicitly stated otherwise, the reuse threshold is 30.0 for XAttention \citep{xu2025xattention} on the Wan models and 8.0 for SVG2 \citep{yang2026sparse}; the HunyuanVideo SVG2 configuration uses 30.0. The drift is the sum of the L1 distances between mean-pooled query and key features, evaluated independently for each head. The layer-level gating thresholds are 0.4 and 0.8: a layer is fully reused below a 0.4 refreshed-head ratio, fully refreshed above 0.8, and otherwise follows the head-wise decisions.

\paragraph{EBC.}

\noindent\textbf{Calibration protocol.}
We calibrate each model--backbone pair independently. For every head, a calibration run evaluates the candidate top-$p$ thresholds \(\mathcal{T}=\{0.85,0.90,0.95\}\) on its assigned prompt and four denoising timesteps; all other heads use the dense value $p=1.0$ during each isolated measurement. For each candidate threshold, the four measurements from the sampled timesteps are averaged to obtain $E_{l,h,k}$ and $S_{l,h,k}$. Since there are three candidate thresholds and four sampled timesteps, each head requires $3\times4=12$ isolated sparse-head probes. Thus, a complete calibration contains $12LH$ probes, where $L$ and $H$ are the number of transformer layers and attention heads. We randomly select one shared set of ten prompts from VBench and use the same set for all model--backbone pairs.

The calibration runner constructs and caches the full dense denoising trajectory for each prompt once, then reuses the cached latent, scheduler timestep, and dense velocity in all isolated-head probes.

To cover the denoising trajectory, we divide the 50 denoising steps into four contiguous intervals, \([0,11]\), \([12,24]\), \([25,36]\), and \([37,49]\), and sample one step uniformly from each interval. This stratification reflects that a head's behavior typically changes gradually between nearby denoising steps, while it can change more substantially across distant steps. The resulting four prompt--timestep measurements are averaged for each candidate threshold.

\noindent\textbf{Frequency-aware velocity error.}
For the 3D FFT of the sparse-induced velocity error, we apply \texttt{fftshift}. Along each transformed axis, frequencies within a radius of \(\lfloor n/4\rfloor\) from the zero-frequency center are labeled low, and the remaining frequencies are labeled high, where $n$ is the length of that axis. We then group coefficients according to their temporal and spatial frequency status: LL contains low temporal frequency and low frequency on both spatial axes; LH contains low temporal frequency and high frequency on at least one spatial axis; HL contains high temporal frequency and low frequency on both spatial axes; and HH contains high temporal frequency and high frequency on at least one spatial axis. The band weights are
\[
(w_{\mathrm{LL}},w_{\mathrm{LH}},w_{\mathrm{HL}},w_{\mathrm{HH}})=(1.0,0.5,0.01,0.01).
\]
They were selected from the controlled frequency-perturbation study in the main paper: low-temporal, low-spatial perturbations cause the largest degradation, followed by LH, while the two high-temporal regions receive substantially smaller weights. The energy normalization in Eq.~(6) uses \(\varepsilon=10^{-10}\).

\noindent\textbf{Optimization and budget.}
We solve the one-threshold-per-head binary ILP with PuLP and its bundled CBC solver. Each ILP selects exactly one candidate for every head. The global constraint is either a minimum realized-sparsity budget or a maximum average threshold, chosen to match the corresponding shared-threshold $p=0.9$ baseline as closely as possible.

\FloatBarrier
\newpage
\raggedbottom
\TheoreticalDetails

\end{document}